\documentclass{article} 
\usepackage[final]{neurips_2025}

\makeatletter
\providecommand\@trackname{}     
\providecommand\@noticestring{}  
\makeatother

\usepackage{amsmath,amsfonts,bm, amsthm}

\newcommand{\pars}[1]{\left( #1 \right)}
\newcommand{\curs}[1]{\left\{ #1 \right\}}
\newcommand{\sqrs}[1]{\left[ #1 \right]}

\newcommand{\norm}[1]{\left\Vert #1 \right\Vert}

\newcommand{\bigo}[1]{\mathcal{O}\pars{#1}}
\newcommand{\mise}{\mathrm{MISE}}
\newcommand{\var}{\mathrm{Var}}
\newcommand{\bias}{\mathrm{Bias}}

\newtheorem{theorem}{Theorem}
\newtheorem{corollary}{Corollary}

\def\eqref#1{equation~\ref{#1}}

\def\1{\bm{1}}

\DeclareMathAlphabet{\mathsfit}{\encodingdefault}{\sfdefault}{m}{sl}
\SetMathAlphabet{\mathsfit}{bold}{\encodingdefault}{\sfdefault}{bx}{n}

\newcommand{\E}{\mathbb{E}}

\usepackage{hyperref}
\usepackage{url}
\usepackage{amssymb}
\usepackage{graphicx}
\usepackage{algorithm}
\usepackage[noend]{algpseudocode}
\usepackage{cleveref}
\usepackage{booktabs}
\usepackage{comment}

\usepackage[utf8]{inputenc} 
\usepackage[T1]{fontenc}    
\usepackage{hyperref}       
\usepackage{url}            
\usepackage{booktabs}       
\usepackage{amsfonts}       
\usepackage{nicefrac}       
\usepackage{microtype}      
\usepackage{xcolor}         

\title{SD-KDE: Score-Debiased Kernel Density Estimation}

\author{%
  Elliot L.\ Epstein$^{1}$\thanks{Equal contribution.} \And
  Rajat Dwaraknath$^{1}$\footnotemark[1] \And
  Thanawat Sornwanee$^{1}$\footnotemark[1] \AND
  John Winnicki$^{1}$\footnotemark[1] \And
  Jerry Weihong Liu$^{1}$\footnotemark[1] %
  \AND
  $^{1}$Stanford University, Stanford, CA\,94305, USA\\
  \texttt{\{epsteine, rajatvd, tsornwanee, winnicki, jwl50\}@stanford.edu}
}

\begin{document}

\maketitle

\begin{abstract}
  We propose a novel method for density estimation that leverages an estimated \underline{s}core function to \underline{d}ebias \underline{k}ernel \underline{d}ensity \underline{e}stimation (SD-KDE).
  In our approach, each data point is adjusted by taking a single step along the score function with a specific choice of step size, followed by standard KDE with a modified bandwidth.
  The step size and modified bandwidth are chosen to remove the leading order bias in the KDE, improving the asymptotic convergence rate.
  Our experiments on synthetic tasks in 1D, 2D and on MNIST, demonstrate that our proposed SD-KDE method significantly reduces the mean integrated squared error compared to the standard Silverman KDE, even with noisy estimates in the score function.
  These results underscore the potential of integrating score-based corrections into nonparametric density estimation.
\end{abstract}

\section{Introduction}

Kernel density estimation (KDE)~\citep{10.1214/aoms/1177728190,Parzen1962OnEO} is a widely used nonparametric method for estimating an unknown probability density function from a finite set of data points.
The classical KDE effectively smooths the data by convolving with a kernel function, such as the Gaussian kernel, and then normalizing the result to obtain a density estimate.
KDE finds application in diverse fields such as anomaly detection, clustering~\citep{10.1007/978-3-642-37456-2_14}, data visualization~\citep{Scott2012MultivariateDE},  nonparametric statistical inference~\citep{guerre2000optimal, zhang2008kernel}, and dynamical systems~\citep{JMLR:v19:16-349}.

The classical KDE suffers from a well-known bias-variance trade-off, controlled by the choice of kernel bandwidth \citep{silverman1986density}.
Larger bandwidths lead to smoother estimates with lower variance but higher bias, while smaller bandwidths yield more variable estimates with lower bias~\citep{10.1214/aoms/1177728190, Parzen1962OnEO}.
This trade-off is particularly damaging in cases with highly variable density functions, where the bias can dominate the estimation error.

Recent advances in score-based generative modeling and diffusion processes have demonstrated the power of using the score function---the gradient of the log-density---to reverse a forward process of noise injection, effectively reconstructing the underlying data distribution \citep{10.5555/3495724.3496298}.
Notably, methods such as score matching \citep{hyvarinen2005estimation} and its deep learning extensions, diffusion models, \citep{song2019generative} provide robust estimates of the score function even in complex, high-dimensional settings, without requiring density estimation.

\begin{figure}[t]
  \centering
  \includegraphics[width=\textwidth]{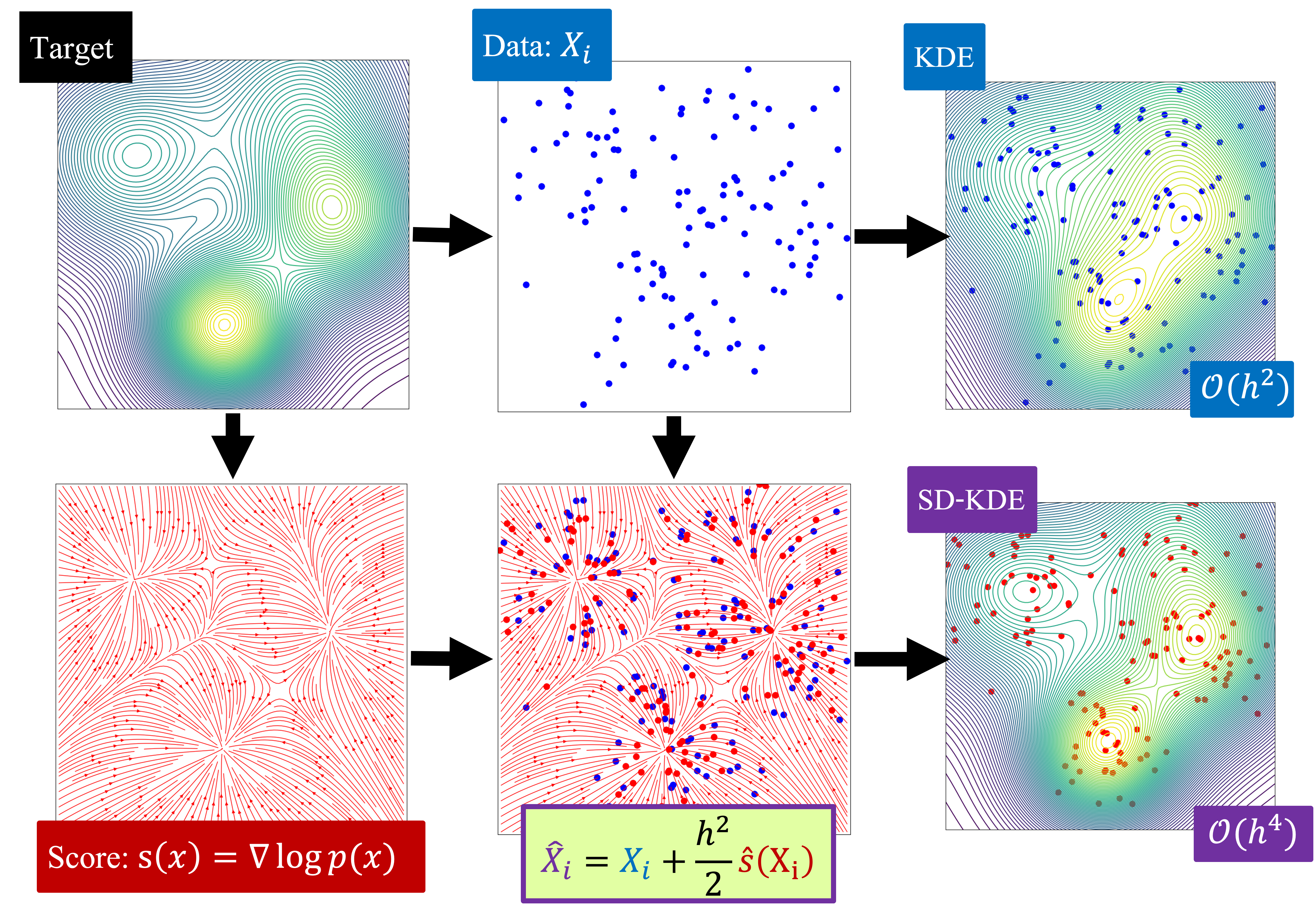}
  \caption{First schematic diagram on SD-KDE. The estimation objective is to estimate the target distribution pdf. In the conventional setting, we have finite samples from the distribution (blue box). Using only this information, we can perform KDE to estimate the probability density function. However, if we have access to the score function, we can combine the data points and score function to get SD-KDE. By fixing the kernel bandwidth to be $h$, we will get that the vanilla KDE and SD-KDE have a pointwise variance of order $\mathcal{O}\left(\frac{1}{nh^d}\right)$. However, SD-KDE reduces the pointwise bias from $\mathcal{O}(h^2)$ to $\mathcal{O}(h^4)$ per the theorem~\ref{thm:onestep_choice}.}
  \label{fig:sdkdewithscore}
\end{figure}

In this work, we investigate whether incorporating knowledge of the score function into the KDE framework allows us to push the Pareto frontier of the bias-variance trade-off. We propose a method to debias the KDE using the score function to improve density estimation accuracy.
Specifically, our method adjusts each data point by taking a small step in the direction of the estimated score, and then performs KDE with a modified bandwidth, as illustrated in Figure~\ref{fig:sdkdewithscore}.
Intuitively, taking a step along the score sharpens the sample distribution, which counteracts the smoothening effect of applying the KDE.
We find that with a carefully chosen combination of step size and KDE bandwidth, we remove the leading order bias in the KDE, resulting in a more accurate, debiased density estimate.


In summary, our contributions are the following:
\begin{enumerate}
    \item We propose \Cref{alg:debiased_kde}, our method for score-debiased kernel density estimation (SD-KDE).
    \item We provide asymptotically optimal bandwidth and step size selection for \Cref{alg:debiased_kde} (\Cref{thm:onestep_choice}), achieving the asymptotic mean integral square error (AMISE) of order $\bigo{n^{-8/(d+8)}}$, instead of the $\bigo{n^{-4/(d+4)}}$ achieved by a standard KDE~\citep{silverman1986density}. 
    \item In \Cref{sec:experiment}, we numerically corroborate our theoretical results on 1D and 2D synthetic datasets and observe strong agreement with the asymptotic scaling identified in \Cref{thm:onestep_choice}.
\end{enumerate}

\section{Method and Theoretical Results}\label{sec:method}


\begin{algorithm}
  \caption{Score-Debiased Kernel Density Estimation}
  \label{alg:debiased_kde}
  \begin{algorithmic}[1] 
    \Require Data $\{x_i\}_{i=1}^n$, score estimator $\hat{s}$, kernel $K$, KDE bandwidth $h$, score step size $\delta$
    \State Take a single step along the score function: $\widetilde{x}_i \gets x_i + \delta \hat{s}(x_i)$ for $i = 1, \ldots, n$
    \State Compute the debiased kernel density estimate: $\hat{p}(x) = \frac{1}{nh^d} \sum_{i=1}^n K\left(\frac{x - \widetilde{x_i}}{h}\right)$
  \end{algorithmic}
\end{algorithm}
\begin{theorem}[Optimal Bandwidth and Step Size selection for $\Cref{alg:debiased_kde}$]
  \label{thm:onestep_choice}
  Let $\curs{x_i}_{i=1}^n$ be i.i.d.
  samples from a smooth density $p$ in $\mathbb{R}^d$.
  Let $\hat{s}$ be the exact score function of $p$.
  Let $K$ be a symmetric kernel with mean $0$ and covariance $\int u u^\top K(u) du = I$.
  The debiased kernel density estimate $\hat{p}$ obtained by running $\Cref{alg:debiased_kde}$ with bandwidth $h$ and step size $\delta$ is given by
  \begin{align*}
    \hat{p}(x) = \frac{1}{nh^d} \sum_{i=1}^n K\pars{\frac{x - \pars{x_i +\delta \hat{s}(x_i)}}{h}}.
  \end{align*}
  The asymptotically optimal bandwidth and step size for $\Cref{alg:debiased_kde}$ are given by
  \begin{align*}
    h_{\text{opt}} = \bigo{n^{-1/(d+8)}}, \quad \delta_{\text{opt}} = \frac{h_{\text{opt}}^2}{2}.
  \end{align*}
  The resulting debiased kernel density estimate $\hat{p}$ satisfies
  \begin{align*}
    \mise := \E\sqrs{\int{\pars{\hat{p}(x) - p(x)}}^2 dx} = \bigo{n^{-8/(d+8)}}.
  \end{align*}
\end{theorem}

We include the detailed proof of \Cref{thm:onestep_choice} in \Cref{app:proof}.

\begin{corollary}
\label{corollary:estimatrescore}
    If the estimate score $\hat{s}(\cdot)$ is not equal to the actual score $s(\cdot)$, the bandwidth is $h$, and the stepsize is $\frac{h^2}{2}$ as in the theorem~\ref{thm:onestep_choice}, then the bias is given by 
    \begin{align*}
        \mathbb{E}\left[\hat{p}(x) - p(x)\right]
        =
        -\frac{h^2}{2}\left[\left(\hat{s}(x)-s(x)\right) \nabla p(x) + p(x)\nabla\left(\hat{s}(x)-s(x)\right)\right] + \mathcal{O}(h^4).
    \end{align*}
\end{corollary}
The proof for this corollary directly follows the proof for the theorem~\ref{thm:onestep_choice}.

\paragraph{Discussion.}

\Cref{thm:onestep_choice} demonstrates that, when a score oracle is available, one can eliminate the asymptotically dominant term, thereby reducing the bias from the conventional order of $\bigo{h^4}$ to $\bigo{h^8}$. Although a higher-order kernel—such as the effective spline kernel described by \citet{silverman1984spline}—can similarly achieve a similar bias reduction, it typically introduces regions where the estimated density assumes negative values. This drawback poses a significant practical challenge, as the numerical normalization of the resulting probability density function is computationally intractable \citep{song2019generative}.

We note that our method flexibly allows a variety of kernels to be used for KDE, since the only requirement for the kernel is in symmetricity and covariance structure, both of which can be conveniently satisfied~\citep{chen2017tutorial}. 

Although \Cref{thm:onestep_choice} requires the knowledge of the score function, we observe empirically that a small discrepancy of the estimated score function and the underlying score function under some level may only have minimal effect on the performance. See \Cref{sec:experiment} for more details. 

\section{Experiments}
\label{sec:experiment}

\subsection{1D Synthetics}

\paragraph{Experimental setup.}
We test the empirical performance of the SD-KDE method on density estimation of 1D Gaussian mixture models, and include a similar analysis for Laplace mixture models in Appendix~\ref{sec:additional_experiments}.
We sample data from three mixtures, where
$$
\label{eq:gaussian_mixture}
p(x) \;=\; \pi \, \mathcal{N}\bigl(x \mid \mu_1, \sigma_1^2\bigr)
\;+\;\bigl(1 - \pi\bigr)\,\mathcal{N}\bigl(x \mid \mu_2, \sigma_2^2\bigr)\,
$$
and each mixture’s parameters \((\pi, \mu_1,\sigma_1,\mu_2,\sigma_2)\) are outlined in Table~\ref{tab:gaussian_mixture_params}.
We compare the SD-KDE method with a baseline based on the classical Silverman KDE, using Silverman's bandwidth formula~\citep{silverman1986density}, given by $
h = 0.9 \cdot \min(\hat{\sigma}, IQR / 1.34) \cdot n^{-1/5}$, where IQR is the inter-quartile range. 
To investigate how sensitive our method is to the estimation accuracy of the score function, we test the performance of our method when only given access to a noisy score function estimate, e.g. we observe $\tilde{S}(x) = S(x) + \epsilon$, where $S(x)$ is the score function and $\epsilon \sim N(0, \sigma)$ for a given standard deviation $\sigma$.
Performance is evaluated with \emph{mean integrated squared error} (MISE).

\label{experiment_runtime} Most of the experiments in the paper were conducted on a Linux cluster with 5 NVIDIA RTX A6000 GPUs, each with 49140 MB memory, running on CUDA Version 12.5. The cluster has 256 AMD EPYC 7763 64-Core Processor CPUs. Some experiments were also conducted on a MacBook Air (2022) equipped with an Apple M2 chip and 16 GB of unified memory. All experiments took less than 1 hour to run. 

\begin{figure}[t]
  \centering
  \includegraphics[width=\textwidth]{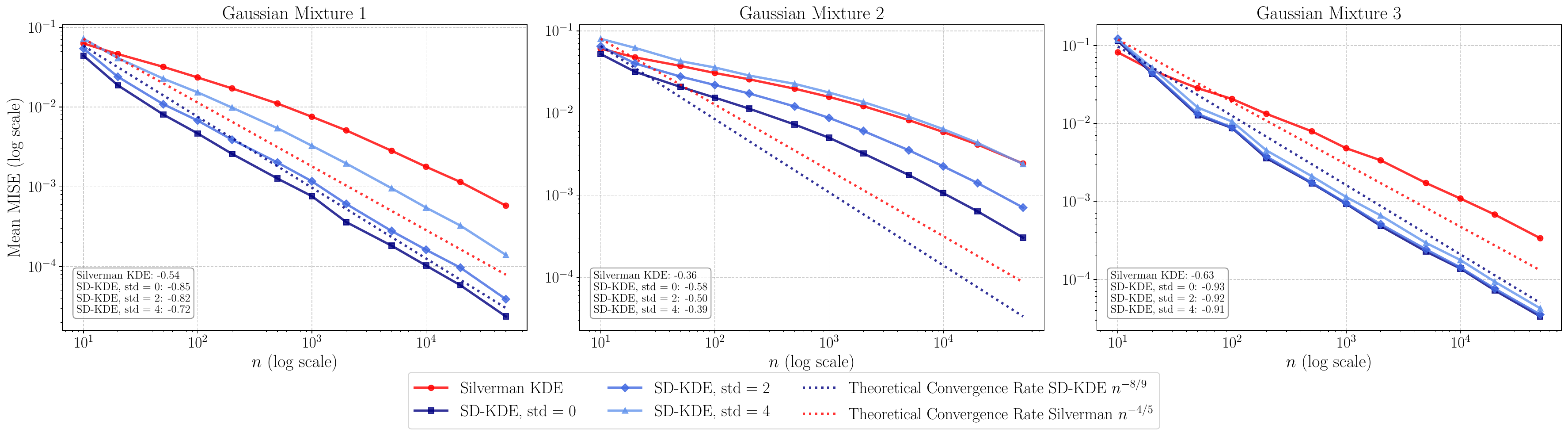}
  \caption{%
    MISE error as a function of $n$ for the three Gaussian mixtures for Silverman vs. SD-KDE. Each point is the average MISE over 50 random seeds. 
    The slopes inside each subplot are fitted regression lines in log--log scale, indicating how quickly each method’s error decays as $n$ increases.
  }
  \label{fig:scaling_experiment_mise}
\end{figure}

\paragraph{SD-KDE is robust to noisy score function estimate.}
In Figure~\ref{fig:scaling_experiment_mise}, we show the MISE of the SD-KDE (as a function of the number of observed samples, $n$), with varying degree of added noise and compare to the Silverman KDE. Each point in the plot represents an average over 50 seeds. We see that the SD-KDE method has a significantly better asymptotic scaling than the Silverman baseline, up to a score function noise level with $\sigma=4$. Even in the presence of a highly noisy score function, we find the SD-KDE method provides a significant gain. We also display the fitted regression slope associated with each line, along with the theoretical asympototic convergence rate of $n^{-8/9}$. 
We note the close tracking between the SD-KDE asymptotic decay ($-0.85$ for mixture 1, and $-0.93$ for mixture 3) compared with the theoretical predicted decay ($=-8/9= -0.86$). For mixture 2, all models have weaker performance due to the challenging mixture shape, indicating that larger $n$ is needed to reach the theoretical decay rate. For $n=5\times10^4$, the SD-KDE has an order of magnitude smaller MISE error on average across 50 seeds compared with the Silverman method.

\begin{figure}[t]
  \centering
  \includegraphics[width=0.8\textwidth]{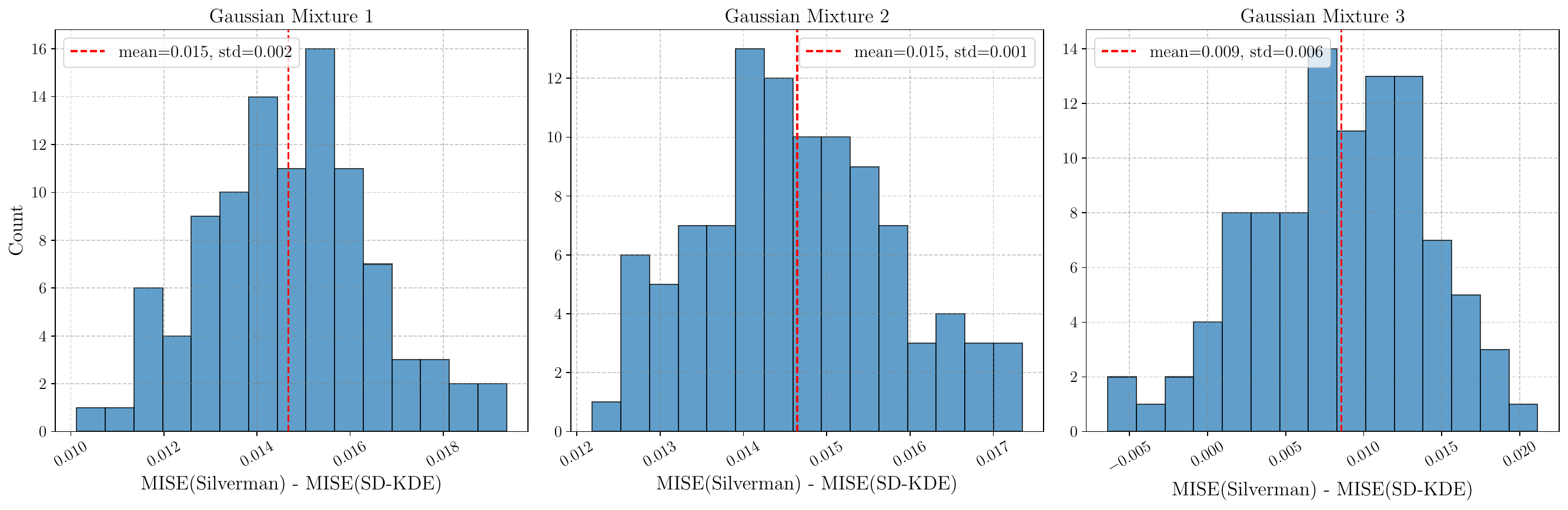}
  \caption{%
    Histogram of MISE difference of the SD-KDE method and the Silverman method, for $n=100$ samples and 50 random seeds per mixture.
    The SD-KDE method is consistently having lower MISE than the Silverman baseline; for mixtures 1 and 2, SD-KDE method outperforms for all 100 samples, and for the third mixture, it is better in 95\% of samples. 
  }
  \label{fig:kl_diff_histograms}
\end{figure}

\begin{figure}[t]
    \centering
    \includegraphics[width=\linewidth]{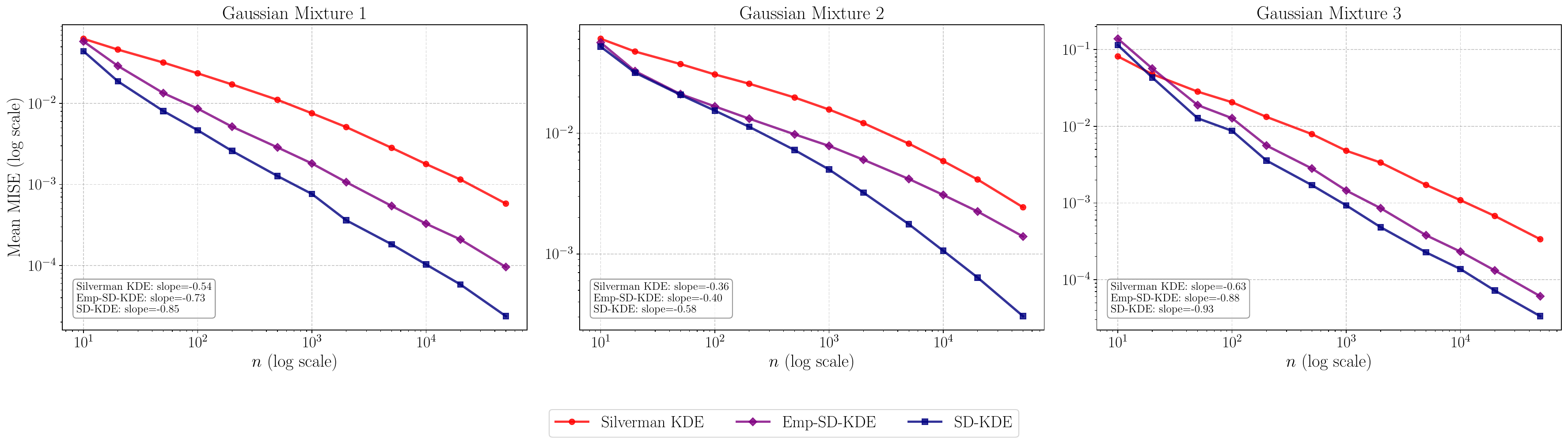}
    \caption{MISE error as a function of $n$ for each of the three gaussian mixtures. For each point, we compute the MISE with 50 random seeds per mixture. 
    Each subplot plots the mean integrated squared error as a function of $n$. 
    The legend compares Silverman KDE, Emp-SD-KDE (estimating the score from the data), and SD-KDE (ground truth score).
    The slopes inside each subplot are fitted regression lines in log--log scale indicating how quickly each method’s error decays as $n$ increases.}
    \label{fig:empirical_sd_kde}
\end{figure}

\paragraph{SD-KDE consistently beats Silverman baseline.}
In Figure~\ref{fig:kl_diff_histograms}, we examine the consistency of the performance gains across multiple data seeds for $n=100$. We observe that the SD-KDE method is consistently better than the Silverman baseline; for mixtures 1 and 2, SD-KDE method outperforms for all 100 samples, and for the third mixture, it is better in 95\% of samples. 

\begin{figure}[t]
  \centering
  \includegraphics[width=\textwidth]{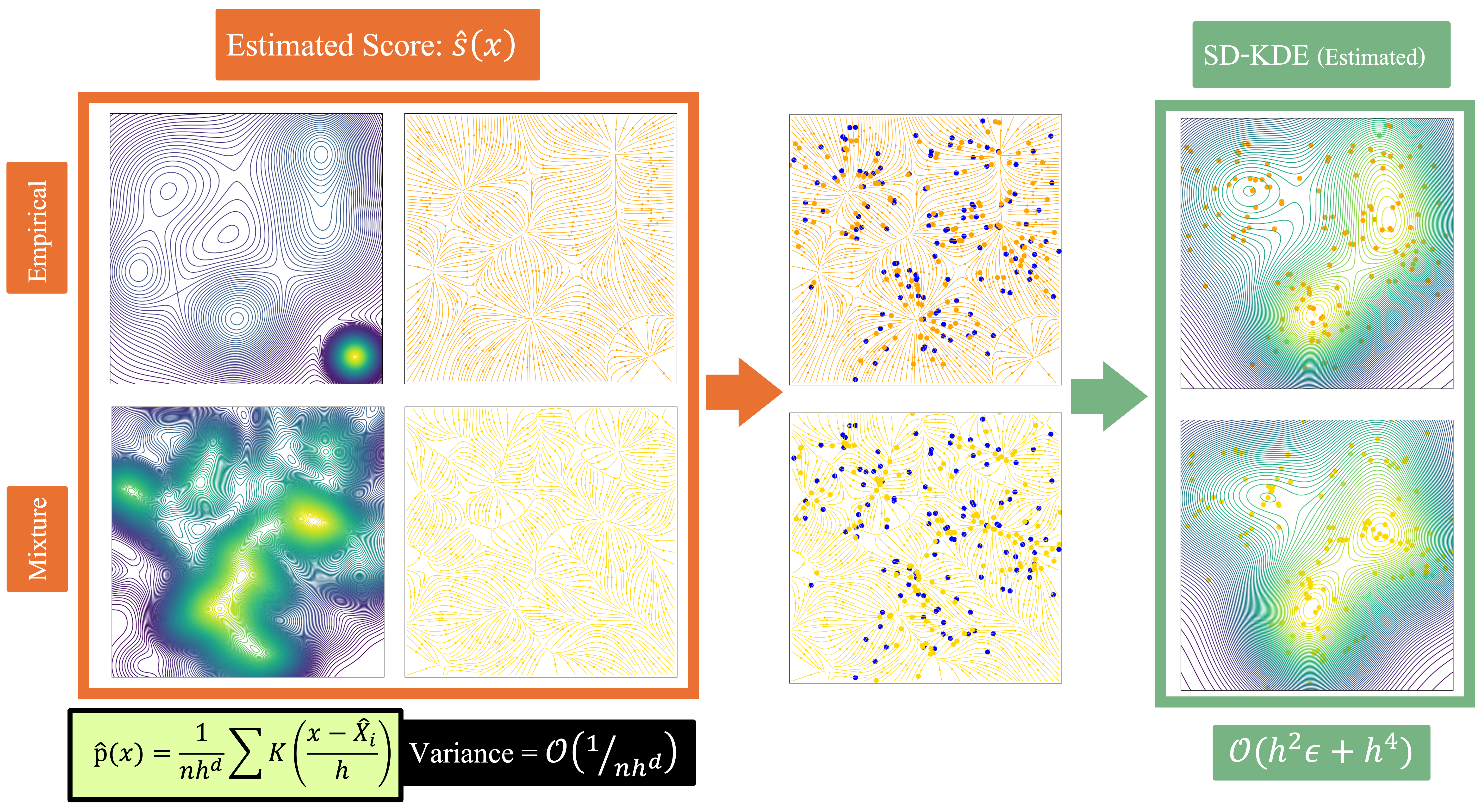}
    \caption{Second schematic diagram on SD-KDE. In the case where a score function is not available. We can use a proxy score function from a proxy distribution. In the example, this is the mixture of the original distribution with some Gaussian distribution. We can also estimate the score from data points. If the estimated score and the actual score are close enough as in the corollary~\ref{corollary:estimatrescore}, then one can attain a better result with SD-KDE compared to vanilla KDE.}
  \label{fig:sdkdewithapproximatescore}
\end{figure}

\paragraph{SD-KDE with the Empirical Score}
We now relax the assumption that the Score function is given, rather, we use Silverman KDE to approximate the score function, and then apply SD-KDE based on this estimated score. We call this method Emp-SD-KDE. Figure~\ref{fig:empirical_sd_kde} shows that Emp-SD-KDE method greatly improves on the standard Silverman KDE, without any assumptions on knowing the true score function. Figure~\ref{fig:sdkdewithapproximatescore} shows how the empirical score is used.

\subsection{2D Synthetics}

\begin{figure}[t]
  \centering
  \includegraphics[width=\textwidth]{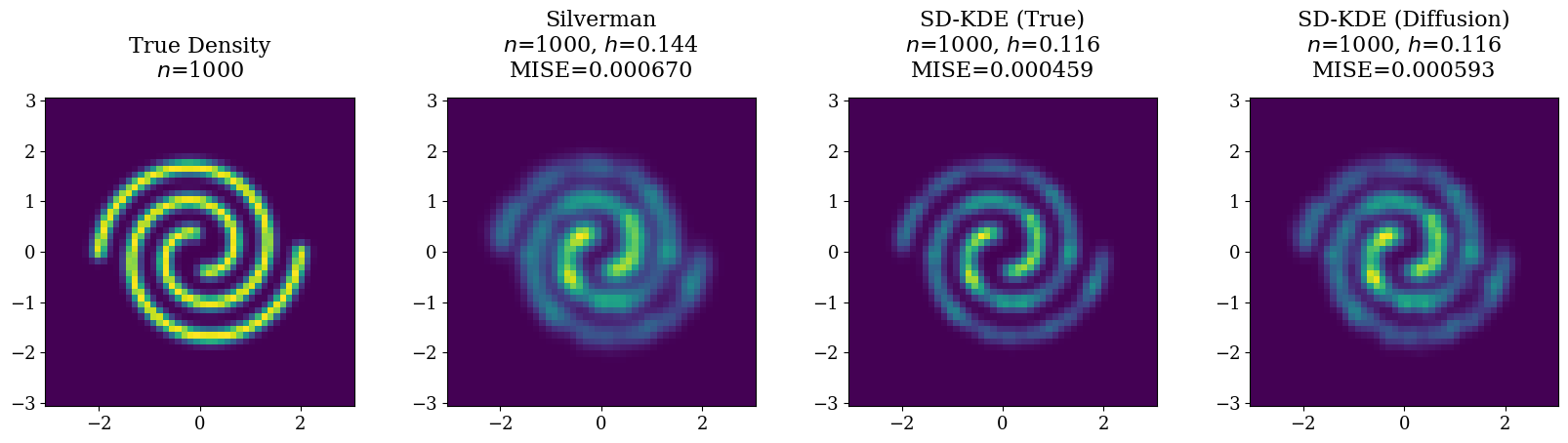}
  \vspace{-.2cm}
  \caption{
    Comparison of the true 2D spiral density vs. Silverman and SD-KDE. For SD-KDE, we evaluate with both the true and learned (diffusion model) score. SD-KDE outperforms Silverman with the oracle score function, and achieves comparable performance even using a noisy score.
  }
  \label{fig:2d_spiral}
\end{figure}

We present preliminary results on 2D synthetic tasks, a spiral distribution (\Cref{fig:2d_spiral}) and a mixture of Gaussians (\Cref{fig:2d_mog}), following \cite{2004.09017, grathwohl2018scalable}.

In \Cref{fig:2d_spiral}, we compare the 2D Silverman method to SD-KDE for the spiral distribution. We compare the accuracy of our method using the true score function to using an estimate of the score function obtained by training a denoising diffusion probabilistic model (DDPM) from scratch on the training data. For the diffusion model architecture, we use a 3-layer MLP with hidden dimension 512, and we train the model with Adam for 1500 steps. We use 1000 diffusion steps during training.
Using the true score function, our proposed method outperforms the Silverman method both qualitatively (via visual assessment) and quantitatively, as measured by the MISE. When employing the score estimated from the diffusion model, our method achieves performance comparable to that of the Silverman method. We attribute this discrepancy with the method under the true score parameter primarily to challenges encountered during the training of the diffusion model rather than to any inherent limitations of the method itself, particularly given the accuracy observed when using the true score.

\subsection{Iterated SD-KDE: Incremental Improvements to KDE}

\begin{figure}[h]
  \centering
  \includegraphics[width=\textwidth]{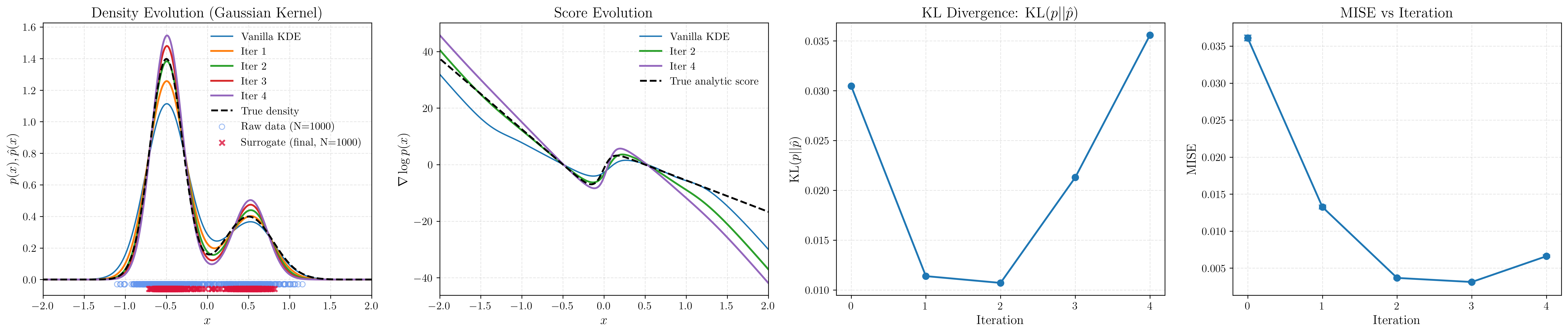}
  \vspace{-.2cm}
  \caption{
    Left to right: (a) Density estimates obtained by vanilla KDE (blue) and by SD‑KDE after one to four score‑debiased iterations (warm colors). The surrogate samples produced by the final iteration (red, $n=1000$) visibly sharpen the bimodal structure relative to the raw data (blue, $n=1000$). (b) The corresponding score functions converge toward the analytic score (black dashed), illustrating progressive removal of higher‑order bias. (c) Kullback–Leibler divergence falls by more than a factor of three after the first correction and attains its minimum at the second iteration before mild over‑correction appears. (d) Monte‑Carlo MISE (mean integrated square error over 200 replicates) mirrors the KL trend, confirming that a small number of SD‑KDE steps yields the best bias–variance trade‑off for this 1D Gaussian mixture. 
  }
  \label{fig:it_gauss}
\end{figure}

In a 1D Gaussian mixture experiment, we examine an iterative application of SD-KDE to further improve the density estimate. We work with a gaussian mixture centered at $\pm 0.5$, each with standard deviation $0.2$, $0.3$ with weights $0.7$ and $0.3$. For this experiment, we will sample $1000$ points and hold the bandwidth constant at $0.15$. We start with a vanilla Gaussian-kernel KDE fit to the mixture data and compute its the closed form solution or approximation of its score, then apply SD-KDE (one score-based correction step, with scale $0.015$ decaying at a rate of $0.15$ at each iteration) to generate surrogate points that remove the leading-order bias. The resulting debiased KDE serves as the baseline for the next iteration, where we recompute the score and apply SD-KDE again; each successive iteration thus leverages a more accurate score estimate to correct residual higher-order biases. Figure~\ref{fig:it_gauss} shows the method when one iteration is taken.  Intuitively, since the first SD-KDE step cancels the dominant bias term, subsequent iterations can target smaller remaining discrepancies, progressively aligning the estimated density more closely with the true distribution. As shown in Figure~\ref{fig:it_gauss}, repeated application of SD-KDE yields a closer alignment between the estimated and true probability densities and a corresponding reduction in KL divergence and mean integrated squared error (MISE) with each iteration. Notably, while a single SD-KDE iteration often captures the majority of the improvement in simpler mixture scenarios (additional iterations confer negligible benefit), more complex multi-modal cases or smaller sample regimes benefit from multiple iterations, albeit with diminishing returns. These results illustrate how SD-KDE could be used to directly improve upon KDE without training a separate score oracle (which can often be difficult to train). Similar to the previous sections, we include a similar analysis for Laplace mixture models in Appendix~\ref{sec:additional_experiments}.

\begin{table}[ht!]
    \centering
    \caption{Parameters for the three univariate Gaussian mixtures used in our experiments. 
             Each mixture follows the generic form 
             $p(x) = \pi\,\mathcal{N}(x \mid \mu_1, \sigma_1^2) + (1-\pi)\,\mathcal{N}(x \mid \mu_2, \sigma_2^2)$.}
    \label{tab:gaussian_mixture_params}
    \vspace{.2cm}
    \begin{tabular}{lcccccc}
    \toprule
    \textbf{Mixture} & ${\pi}$ & ${\mu_1}$ & ${\sigma_1}$ 
                     & ${\mu_2}$ & ${\sigma_2}$ \\
    \midrule
    \textbf{1} & 0.4 & -2.0 & 0.5 & 2.0 & 1.0 \\
    \textbf{2} & 0.3 & -2.0 & 0.4 & 4.0 & 1.5 \\
    \textbf{3} & 0.5 & 0.0  & 0.4 & 1.5 & 1.5 \\
    \bottomrule
    \end{tabular}
\end{table}
\subsection{MNIST Dataset}
In this study, we follow a similar experimental setup to \cite{2004.09017} and explore the relationship between generated image quality and estimated density using the MNIST dataset—a widely recognized benchmark comprising 70,000 grayscale images ($28 \times 28$ pixels) of handwritten digits \citep{lecun-mnisthandwrittendigit-2010}. We trained a DDPM on this dataset and, by selecting the lowest diffusion timestep ($t=1$), obtained an estimate of the score function for individual images. Using this score, we apply SD-KDE in latent space to assess the realism of generated images. We ranked generated images from highest to lowest estimated probability density, visualized in \Cref{fig:numbers}. The images with higher density appear more realistic and are correlated with higher quality.

\section{Proof of \Cref{thm:onestep_choice}}\label{app:proof}

\begin{proof}
  First, we decompose the MISE into the bias and the variance terms as 
  \begin{align*}
    \mise & = \int \pars{\E\sqrs{\hat{p}(x)} - p(x)}^2 dx+ \int \pars{\E\sqrs{\hat{p}(x)^2} - \E \sqrs{\hat{p}(x)}^2} dx  \\
          & =  \int \bias\sqrs{\hat{p}(x)}^2 dx + \int \var\sqrs{\hat{p}(x)} dx,
  \end{align*}
  where the variance term
  $\var\sqrs{\hat{p}(x)} = \E\sqrs{\hat{p}(x)^2} - \E \sqrs{\hat{p}(x)}^2$ and the bias term $\bias\sqrs{\hat{p}(x)} = p(x) - \E\sqrs{\hat{p}(x)}$.

  The variance term is given by
  \begin{align*}
    \var\sqrs{\hat{p}(x)} = \frac{1}{n} \var \pars{\frac{1}{h^d}
      K\pars{\frac{x - \pars{X + \delta \hat{s}(X)}}{h}}}
  \end{align*}
  since $\hat{p}(x)$ is a sum of $n$ i.i.d. terms. Using Taylor expansion at the kernel $K$ around $\frac{x-X}{h}$ yields
  \begin{align*}
    K\left( \frac{x - (X + \delta \hat{s}(X))}{h} \right)
    = K\left( \frac{x - X}{h} \right) - \frac{\delta}{h} \hat{s}(X)^\top \nabla K\left( \frac{x - X}{h} \right) + O\left( \frac{\delta^2}{h^2} \right).
  \end{align*}

  The variance is dominated by the leading order term, which gives
  \begin{align*}
    \var\sqrs{\hat{p}(x)}  = \frac{1}{n} \var \pars{\frac{1}{h^d}
    K\pars{\frac{x - X}{h}}} + \bigo{\frac{\delta^2}{nh^{2+d}}}= \bigo{\frac{1}{nh^d}+\frac{\delta^2}{nh^{d+2}}}
  \end{align*}
where we used the standard KDE variance result for the leading term.

  Now, we analyze the bias term.
  \[
    \text{Bias}[\hat{p}(x)] = \mathbb{E}[\hat{p}(x)] - p(x).
  \]
  We write the expectation of $\hat{p}(x)$ as
  \begin{align*}
    \E\sqrs{\hat{p}(x)}  = \frac{1}{h^d} \E\sqrs{K\pars{\frac{x - \pars{X + \delta \hat{s}(X)}}{h}}} = \int \frac{1}{h^d}
    K\pars{\frac{x - \pars{y + \delta \hat{s}(y)}}{h}} p(y) dy.
  \end{align*}
  We substitute $u = \frac{x - y}{h}$ to obtain
  \begin{align}
    \label{eq:bias}
    \E\sqrs{\hat{p}(x)} & = \int K\pars{u - \frac{\delta}{h} \hat{s}(x - hu)} p(x - h u) du.
  \end{align}

  Taylor expansion will yield that
  \begin{align*}
    p(x - h u) = p(x) - h u^\top \nabla p(x) + \frac{h^2}{2} u^\top \nabla^2 p(x) u +
    \bigo{h^3},
  \end{align*}
  and that
  \begin{align*}
    &K\pars{u - \frac{\delta}{h} \hat{s}(x - hu)}\\ &=  
    K\pars{u - \frac{\delta}{h} \hat{s}(x) + \delta u^\top \nabla \hat{s}(x) + \bigo{\delta h}}\\
    &=
    K(u) - \frac{\delta}{h} \hat{s}(x)^\top \nabla K(u) + \delta u^T \nabla \hat{s}(x) \nabla K(u) 
     + \frac{\delta^2}{2h^2} \hat{s}(x)^\top \nabla^2 K(u) \hat{s}(x) + \bigo{\delta h + \frac{\delta^2
    }{h} + \frac{\delta^3}{h^3}}.
  \end{align*}

  Substitute these expansions into \Cref{eq:bias}, and expand the product.
  We consider each term separately.

  \begin{enumerate}
    \item $K(u) p(x)$ integrates to $p(x)$ by the definition of $K$.
    \item $- \frac{\delta}{h} \hat{s}(x)^\top \nabla K(u) p(x)$ integrates to $0$ since $K$ is symmetric and decays to $0$ at infinity.
    \item $K(u) \pars{- h u^\top \nabla p(x)}$ integrates to $0$ by the symmetry of $K$.
    \item $K(u) \pars{\frac{h^2}{2}u^\top \nabla^2 p(x) u}$ integrates to $\frac{h^2}{2} \text{Tr}\pars{\nabla^2 p(x) \int uu^\top K(u) du} = \frac{h^2}{2} \nabla^2 p(x)$.
    \item $- \frac{\delta}{h} \hat{s}(x)^\top \nabla K (u) \cdot (-h u^\top \nabla p(x))$. Integrate by parts on $\nabla K(u)$ to obtain
          \begin{align*}
            \delta \hat{s}(x)^\top \int u^\top \nabla p(x) \nabla K(u) du  = \delta \hat{s}(x)^\top \pars{- \int K(u) \nabla p(x) du}  = - \delta \hat{s}(x)^\top \nabla p(x).
          \end{align*}
          Using $\hat{s}(x) = \nabla \log p(x)$, we have
          \begin{align*}
            - \delta \hat{s}(x)^\top \nabla p(x) = -\delta \nabla \log p(x)^\top \nabla p(x) = - \delta \frac{\norm{\nabla p(x)}^2}{p(x)}.
          \end{align*}
          Using a standard multivariable calculus identity, we have
          \begin{align*}
            -\delta \frac{\norm{\nabla p(x)}^2}{p(x)} = -\delta (\nabla^2 p(x) - p(x) \nabla^2 (\log p(x))).
          \end{align*}
    \item $p(x) \delta u^T \nabla \hat{s}(x) \nabla K(u)$. Again, after integration by parts, we obtain
          \begin{align*}
            \delta \int u^T \nabla \hat{s}(x) \nabla K(u) p(x) du  = \delta \int p(x) K(u) \text{tr}\pars{\nabla \hat{s}(x)} du = \delta p(x) \text{tr}\pars{\nabla \hat{s}(x)} =\delta p(x) \nabla^2 (\log p(x)).
          \end{align*}
    \item $p(x) \frac{\delta^2}{2h^2} \hat{s}(x)^\top \nabla^2 K(u) \hat{s}(x)$ integrates to $0$.
  \end{enumerate}
  Using smoothness, we then have that
  \begin{align*}
      \E\sqrs{\hat{p}(x)} - p(x)
      =
      \frac{h^2}{2}\nabla^2 p(x) -\delta \nabla^2 p(x)+
      \bigo{h^3+\delta h + \frac{\delta^2}{h} + \frac{\delta^3}{h^3}}
  \end{align*}
  Now, by choosing $\delta = \frac{h^2}{2}$, we make the leading term zero, and the bias $\E\sqrs{\hat{p}(x)} - p(x) = \bigo{h^3+\delta h + \frac{\delta^2}{h} + \frac{\delta^3}{h^3}} = \bigo{h^3}$.
  Using the standard KDE argument (symmetry of $K$ and decay to $0$ at infinity), we can show that $h^3$ terms in the bias also vanish.
  Thus, the bias is $\bigo{h^4}$.

  Moreover, note that $\delta = \frac{h^2}{2}$, so the variance term $\var\sqrs{\hat{p}(x)} = \bigo{\frac{1}{nh^d}}$.

  For optimal error scaling, we balance the bias and the leading variance terms.
  The error due to bias is $\bigo{h^8}$, and the leading error due to variance is $\bigo{\frac{1}{nh^d}}$.

  Balancing these terms, we obtain $h_{\text{opt}} = \bigo{n^{-1/(d+8)}}$.

  Finally, the $\mise$ is $\bigo{h^8} = \bigo{n^{-8/(d+8)}}$.
\end{proof}

\section{Additional Discussion}
\paragraph{Connections to Langevin dynamics.}
We note that the algorithm is an analog to the continuous time Langevin dynamics, which uses the score function $s$ and yields that the stochastic differential equation 
\begin{align}
    dX_t = \frac{1}{2} s\left(X_t\right)dt +dB_t
\end{align}
will have the stationary distribution according to the probability distribution function $p$, which corresponds to the score function $s$~\citep{song2019generative, song2020score}. Our work can be viewed as a one-step Euler–Maruyama discretization of the Langevin dynamics to estimate the location-shifted kernel from the sample points. This ensures both tractability as well as the benefit of bias-reduction as seen in the main theorem (\Cref{thm:onestep_choice}). To our knowledge, this is the first approach that employs Langevin dynamics to inform a position-based debiased kernel density estimator. 

\paragraph{Bridging Score-Based and Sample-Based Density Estimation.}
    
    While the paper~\citep{song2020score} suggests a formulation of the flow ODE as an evolution of the density function from an approximate posterior. This approach is prior-free and the flow maps a scaled Gaussian distribution to the data distribution. However, this process does not utilize the availability of samples and relies solely on the score estimate. Since this scheme requires spatial and temporal discretization for density estimation, it is computationally less feasible due to the curse of dimensionality.

    Many works in non-parametric methods (ie. KDE, histogram)~\citep{ silverman1986density,10.1214/aoms/1177728190, Parzen1962OnEO, scott1979optimal, lugosi1996consistency} and neural-based density estimation~\citep{doi:10.1073/pnas.2101344118, magdon1998neural, rezende2015variational, Dinh2016DensityEU, berg2018sylvester} use the sample points for the density estimation, but do not incorporate score function in the density estimation framework. 


    A promising future direction is to consider a multi-step discretization of the Langevin dynamics to obtain asymptotically superior debiasing. Using higher order discretization schemes is also an interesting avenue that we are currently exploring. The multi-step approach introduces more challenges, including non-Gaussianity of the final kernel, since it will be a convolution of multiple Gaussian kernels with different score-dependent shifts.

\vspace{-0.3cm}
\section{Conclusion}
\vspace{-0.2cm}
\label{sec:conclusion}
In this work, we demonstrate that incorporating score information can asymptotically improve density estimation accuracy. We propose a method for score-debiased kernel density estimation that achieves $\bigo{n^{-8/(d+8)}}$ convergence rate in mean integrated squared error, improving upon the classical $\bigo{n^{-4/(d+4)}}$ rate of standard KDE. Our experiments on a variety of synthetic datasets validate these theoretical predictions and show that the method remains effective even when using noisy score estimates, suggesting practical applicability beyond settings where the true score is known.
\paragraph{Limitations.} \label{limitations} A key limitation of our proposed method is that the theoretical performance guarantees for SD-KDE require access to an exact score oracle, which is typically unavailable in practical scenarios. Although our empirical results demonstrate that accurate score estimates obtained from state-of-the-art methods (such as score matching or diffusion models) still provide significant performance improvements, these estimation methods themselves can be computationally expensive, particularly in high-dimensional settings or with large datasets. Future work might explore more efficient score estimation techniques or approximate methods that retain the benefits of SD-KDE while reducing the associated computational overhead. 

\bibliography{iclr2025_conference}

\begin{thebibliography}{24}
\providecommand{\natexlab}[1]{#1}
\providecommand{\url}[1]{\texttt{#1}}
\expandafter\ifx\csname urlstyle\endcsname\relax
  \providecommand{\doi}[1]{doi: #1}\else
  \providecommand{\doi}{doi: \begingroup \urlstyle{rm}\Url}\fi

\bibitem[Berg et~al.(2018)Berg, Hasenclever, Tomczak, and Welling]{berg2018sylvester}
Rianne van~den Berg, Leonard Hasenclever, Jakub~M Tomczak, and Max Welling.
\newblock Sylvester normalizing flows for variational inference.
\newblock \emph{arXiv preprint arXiv:1803.05649}, 2018.

\bibitem[Campello et~al.(2013)Campello, Moulavi, and Sander]{10.1007/978-3-642-37456-2_14}
Ricardo J. G.~B. Campello, Davoud Moulavi, and Joerg Sander.
\newblock Density-based clustering based on hierarchical density estimates.
\newblock In Jian Pei, Vincent~S. Tseng, Longbing Cao, Hiroshi Motoda, and Guandong Xu (eds.), \emph{Advances in Knowledge Discovery and Data Mining}, pp.\  160--172, Berlin, Heidelberg, 2013. Springer Berlin Heidelberg.
\newblock ISBN 978-3-642-37456-2.

\bibitem[Chen(2017)]{chen2017tutorial}
Yen-Chi Chen.
\newblock A tutorial on kernel density estimation and recent advances.
\newblock \emph{Biostatistics \& Epidemiology}, 1\penalty0 (1):\penalty0 161--187, 2017.

\bibitem[Dinh et~al.(2016)Dinh, Sohl-Dickstein, and Bengio]{Dinh2016DensityEU}
Laurent Dinh, Jascha~Narain Sohl-Dickstein, and Samy Bengio.
\newblock Density estimation using real nvp.
\newblock \emph{ArXiv}, abs/1605.08803, 2016.
\newblock URL \url{https://api.semanticscholar.org/CorpusID:8768364}.

\bibitem[Grathwohl et~al.(2019)Grathwohl, Chen, Bettencourt, and Duvenaud]{grathwohl2018scalable}
Will Grathwohl, Ricky T.~Q. Chen, Jesse Bettencourt, and David Duvenaud.
\newblock Scalable reversible generative models with free-form continuous dynamics.
\newblock In \emph{International Conference on Learning Representations}, 2019.
\newblock URL \url{https://openreview.net/forum?id=rJxgknCcK7}.

\bibitem[Guerre et~al.(2000)Guerre, Perrigne, and Vuong]{guerre2000optimal}
Emmanuel Guerre, Isabelle Perrigne, and Quang Vuong.
\newblock Optimal nonparametric estimation of first-price auctions.
\newblock \emph{Econometrica}, 68\penalty0 (3):\penalty0 525--574, 2000.

\bibitem[Hang et~al.(2018)Hang, Steinwart, Feng, and Suykens]{JMLR:v19:16-349}
Hanyuan Hang, Ingo Steinwart, Yunlong Feng, and Johan~A.K. Suykens.
\newblock Kernel density estimation for dynamical systems.
\newblock \emph{Journal of Machine Learning Research}, 19\penalty0 (35):\penalty0 1--49, 2018.
\newblock URL \url{http://jmlr.org/papers/v19/16-349.html}.

\bibitem[Ho et~al.(2020)Ho, Jain, and Abbeel]{10.5555/3495724.3496298}
Jonathan Ho, Ajay Jain, and Pieter Abbeel.
\newblock Denoising diffusion probabilistic models.
\newblock In \emph{Proceedings of the 34th International Conference on Neural Information Processing Systems}, NIPS '20, Red Hook, NY, USA, 2020. Curran Associates Inc.
\newblock ISBN 9781713829546.

\bibitem[Hyv{\"a}rinen \& Dayan(2005)Hyv{\"a}rinen and Dayan]{hyvarinen2005estimation}
Aapo Hyv{\"a}rinen and Peter Dayan.
\newblock Estimation of non-normalized statistical models by score matching.
\newblock \emph{Journal of Machine Learning Research}, 6\penalty0 (4), 2005.

\bibitem[LeCun \& Cortes(2010)LeCun and Cortes]{lecun-mnisthandwrittendigit-2010}
Yann LeCun and Corinna Cortes.
\newblock {MNIST} handwritten digit database.
\newblock 2010.
\newblock URL \url{http://yann.lecun.com/exdb/mnist/}.

\bibitem[Liu et~al.(2020)Liu, Xu, Jiang, and Wong]{2004.09017}
Qiao Liu, Jiaze Xu, Rui Jiang, and Wing~Hung Wong.
\newblock Roundtrip: A deep generative neural density estimator.
\newblock 2020.
\newblock \doi{10.1073/pnas.2101344118}.

\bibitem[Liu et~al.(2021)Liu, Xu, Jiang, and Wong]{doi:10.1073/pnas.2101344118}
Qiao Liu, Jiaze Xu, Rui Jiang, and Wing~Hung Wong.
\newblock Density estimation using deep generative neural networks.
\newblock \emph{Proceedings of the National Academy of Sciences}, 118\penalty0 (15):\penalty0 e2101344118, 2021.
\newblock \doi{10.1073/pnas.2101344118}.
\newblock URL \url{https://www.pnas.org/doi/abs/10.1073/pnas.2101344118}.

\bibitem[Lugosi \& Nobel(1996)Lugosi and Nobel]{lugosi1996consistency}
G{\'a}bor Lugosi and Andrew Nobel.
\newblock Consistency of data-driven histogram methods for density estimation and classification.
\newblock \emph{The Annals of Statistics}, 24\penalty0 (2):\penalty0 687--706, 1996.

\bibitem[Magdon-Ismail \& Atiya(1998)Magdon-Ismail and Atiya]{magdon1998neural}
Malik Magdon-Ismail and Amir Atiya.
\newblock Neural networks for density estimation.
\newblock \emph{Advances in Neural Information Processing Systems}, 11, 1998.

\bibitem[Parzen(1962)]{Parzen1962OnEO}
Emanuel Parzen.
\newblock On estimation of a probability density function and mode.
\newblock \emph{Annals of Mathematical Statistics}, 33:\penalty0 1065--1076, 1962.
\newblock URL \url{https://api.semanticscholar.org/CorpusID:122932724}.

\bibitem[Rezende \& Mohamed(2015)Rezende and Mohamed]{rezende2015variational}
Danilo Rezende and Shakir Mohamed.
\newblock Variational inference with normalizing flows.
\newblock In \emph{International conference on machine learning}, pp.\  1530--1538. PMLR, 2015.

\bibitem[Rosenblatt(1956)]{10.1214/aoms/1177728190}
Murray Rosenblatt.
\newblock {Remarks on Some Nonparametric Estimates of a Density Function}.
\newblock \emph{The Annals of Mathematical Statistics}, 27\penalty0 (3):\penalty0 832 -- 837, 1956.
\newblock \doi{10.1214/aoms/1177728190}.
\newblock URL \url{https://doi.org/10.1214/aoms/1177728190}.

\bibitem[Scott(1979)]{scott1979optimal}
David~W Scott.
\newblock On optimal and data-based histograms.
\newblock \emph{Biometrika}, 66\penalty0 (3):\penalty0 605--610, 1979.

\bibitem[Scott(2012)]{Scott2012MultivariateDE}
David~W. Scott.
\newblock Multivariate density estimation and visualization.
\newblock 2012.
\newblock URL \url{https://api.semanticscholar.org/CorpusID:1253508}.

\bibitem[Silverman(1984)]{silverman1984spline}
Bernard~W Silverman.
\newblock Spline smoothing: the equivalent variable kernel method.
\newblock \emph{The annals of Statistics}, pp.\  898--916, 1984.

\bibitem[Silverman(1986)]{silverman1986density}
Bernard~W Silverman.
\newblock \emph{Density Estimation for Statistics and Data Analysis}, volume~26.
\newblock CRC Press, 1986.

\bibitem[Song \& Ermon(2019)Song and Ermon]{song2019generative}
Yang Song and Stefano Ermon.
\newblock Generative modeling by estimating gradients of the data distribution.
\newblock \emph{Advances in neural information processing systems}, 32, 2019.

\bibitem[Song et~al.(2020)Song, Sohl-Dickstein, Kingma, Kumar, Ermon, and Poole]{song2020score}
Yang Song, Jascha Sohl-Dickstein, Diederik~P Kingma, Abhishek Kumar, Stefano Ermon, and Ben Poole.
\newblock Score-based generative modeling through stochastic differential equations.
\newblock \emph{arXiv preprint arXiv:2011.13456}, 2020.

\bibitem[Zhang et~al.(2008)Zhang, Tian, and Zhang]{zhang2008kernel}
Dongling Zhang, Yingjie Tian, and Peng Zhang.
\newblock Kernel-based nonparametric regression method.
\newblock In \emph{2008 IEEE/WIC/ACM International Conference on Web Intelligence and Intelligent Agent Technology}, volume~3, pp.\  410--413. IEEE, 2008.

\end{thebibliography}
\bibliographystyle{iclr2025_conference}

\newpage

\appendix
\section{Synthetic 1D experiments}
\label{sec:additional_experiments}
\begin{figure}[ht]
  \centering
  \includegraphics[width=0.9\textwidth]{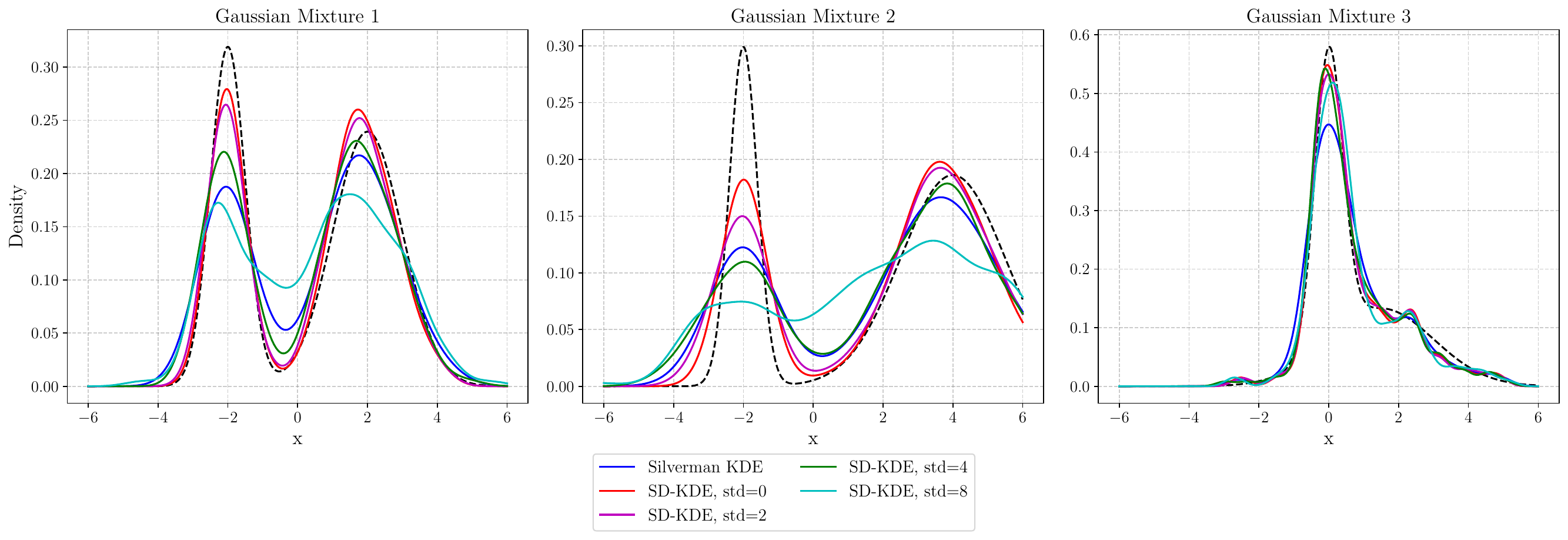}
  \caption{%
    Drawing $n=200$ samples from each of the three Gaussian mixtures in \eqref{eq:gaussian_mixture}
    The dashed black line is the \emph{true} PDF, while the colored lines represent the estimated PDFs.
  }
  \label{fig:example_comparison_kdes}
\end{figure}

\begin{figure}[ht!]
  \centering
  \includegraphics[width=0.8\textwidth]{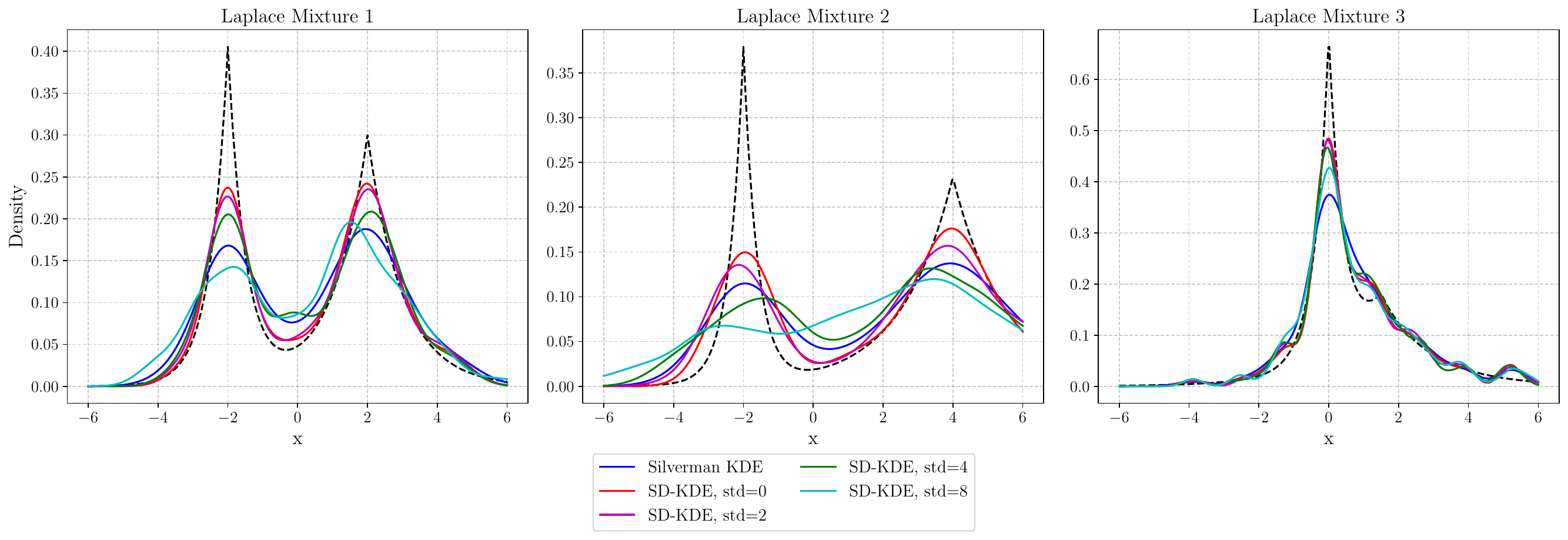}
  \caption{%
    Drawing $n=200$ samples from each of the three Laplace mixtures in \eqref{eq:gaussian_mixture}
    The dashed black line is the true probability density function, while the colored lines represent the estimated probability density functions.
  }
  \label{fig:example_comparison_kdes_laplace}
\end{figure}

\begin{figure}[ht!]
  \centering
  \includegraphics[width=0.8\textwidth]{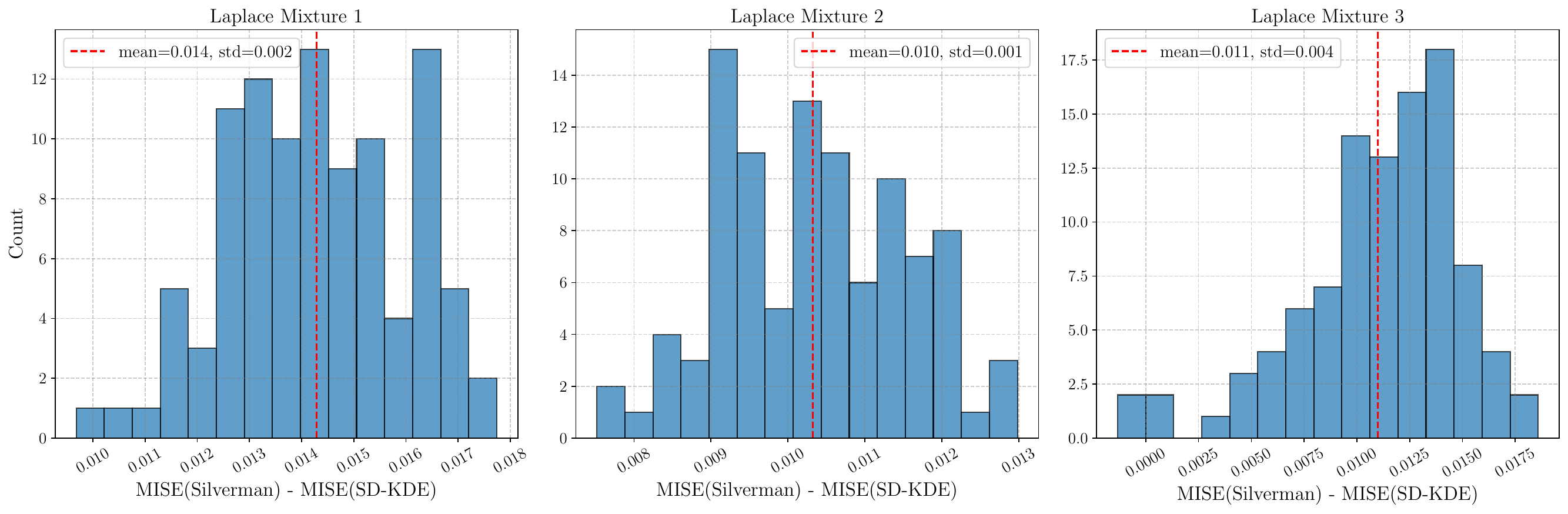}
  \caption{%
    Histogram of MISE difference of the SD-KDE method and the Silverman method, for $n=100$ samples and 50 random seeds per mixture.
    A positive value in the plot indicates that the SD-KDE method performed better for that seed.
    We observe that SD-KDE consistently performs better than the Silverman method over multiple random seeds.
  }
  \label{fig:kl_diff_histograms_laplace}
\end{figure}

Figure~\ref{fig:example_comparison_kdes} and Figure~\ref{fig:example_comparison_kdes_laplace} shows the fitted densities for different noise levels of the SD-KDE method, as well as the Silverman baseline, for $n=200$ samples for three different Gaussian (and Laplace respectively) mixture models, with parameters outlined in Table~\ref{tab:gaussian_mixture_params}. 
In Figure~\ref{fig:kl_diff_histograms_laplace}, we examine the consistency of the performance gains for the SD-KDE method over the Silverman baseline for a mixture of Laplace densities. The Laplace mixtures use the same location and scale parameters as the Gaussian Mixture, given in Table~\ref{tab:gaussian_mixture_params}.

\begin{figure}[ht!]
  \centering
  \includegraphics[width=0.9\textwidth]{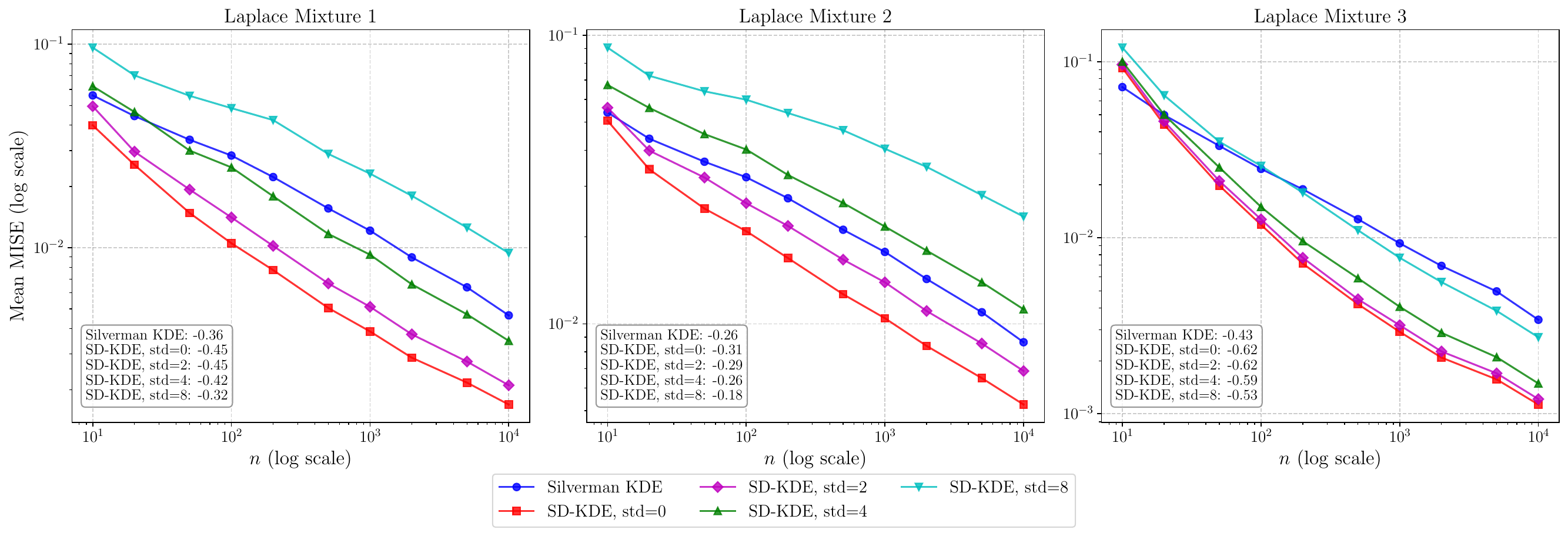}
  \caption{
    MISE error as a function of $n$ for each of the three gaussian mixtures. For each point, we compute the MISE with 50 random seeds per mixture. 
    Each subplot plots the mean integrated squared error as a function of $n$. 
    The legend compares Silverman KDE to SD-KDE at multiple noise settings. 
    The slopes inside each subplot are fitted regression lines in log--log scale indicating how quickly each method’s error decays as $n$ increases.
  }
  \label{fig:scaling_experiment_mise_laplace}
\end{figure}

\begin{figure}[ht!]
  \centering
  \includegraphics[width=0.8\textwidth]{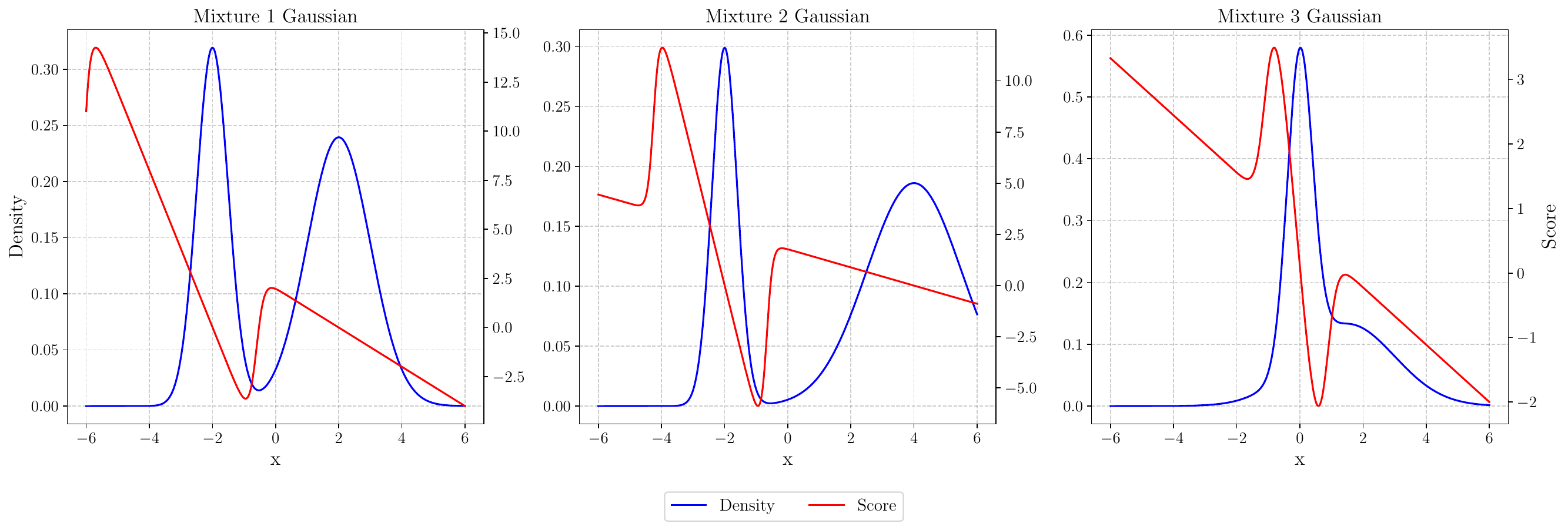}
  \caption{%
    In each subplot, we plot the Gaussian mixture’s density (blue, left axis) and the log-density derivative (score) in red (right axis). 
  }
  \label{fig:score_function_gaussian}
\end{figure}

\begin{figure}[ht!]
  \centering
  \includegraphics[width=0.8\textwidth]{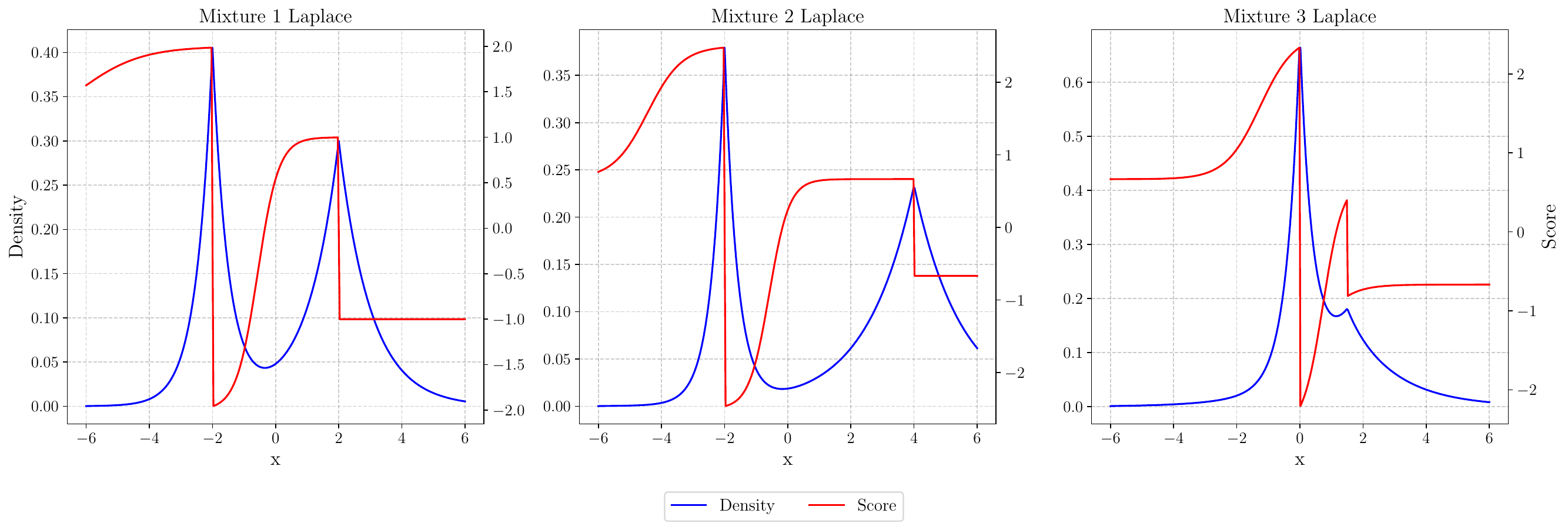}
  \caption{%
    In each subplot, we plot the Laplace mixture’s density (blue, left axis) and the log-density derivative (score) in red (right axis). 
  }
  \label{fig:score_function_laplace}
\end{figure}

Next, we show the scaling in $n$ for a density estimation task for Laplace Mixtures. 
Figure~\ref{fig:scaling_experiment_mise_laplace} shows the results. In Figure~\ref{fig:score_function_gaussian}, and \ref{fig:score_function_laplace}, we show a visualization of the score function and the densities for both the Gaussian and Laplace Mixtures. 

\begin{figure}[t]
  \centering
  \includegraphics[width=\textwidth]{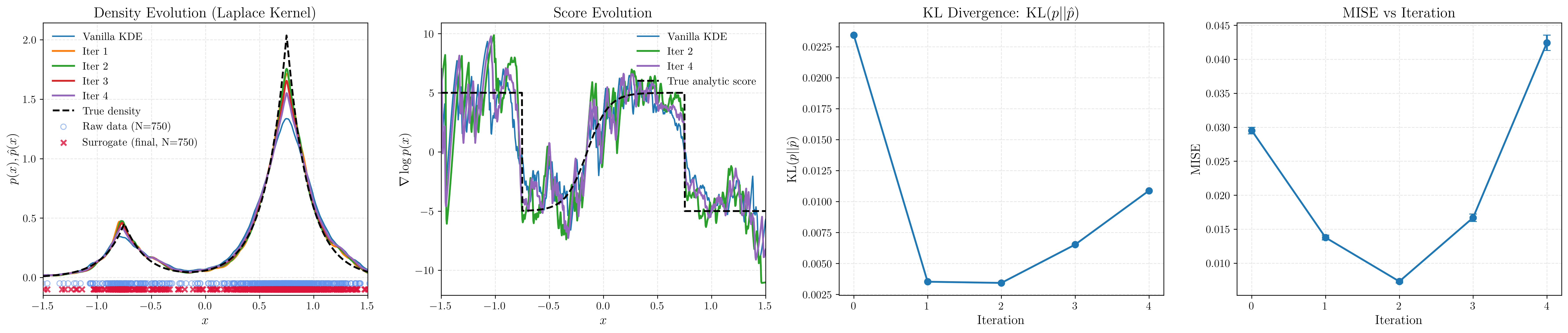}
  \vspace{-.2cm}
  \caption{
    Left to right: (a) Density estimates obtained by vanilla KDE (blue) and by SD‑KDE after one to four score‑debiased iterations (warm colours). The surrogate samples produced by the final iteration (red, $n=1000$) visibly sharpen the bimodal structure relative to the raw data (blue, $n=1000$). (b) The corresponding score functions converge toward the analytic score (black dashed), illustrating progressive removal of higher‑order bias. (c) Kullback–Leibler divergence falls by more than a factor of three after the first correction and attains its minimum at the second iteration before mild over‑correction appears. (d) Monte‑Carlo MISE (mean integrated square error over 200 replicates) mirrors the KL trend, confirming that a small number of SD‑KDE steps yields the best bias–variance trade‑off for this 1D Laplacian mixture. 
  }
  \label{fig:it_laplace}
\end{figure}

\section{Synthetic 2D mixture of Gaussians}
\begin{figure}[ht]
  \centering
  \includegraphics[width=\textwidth]{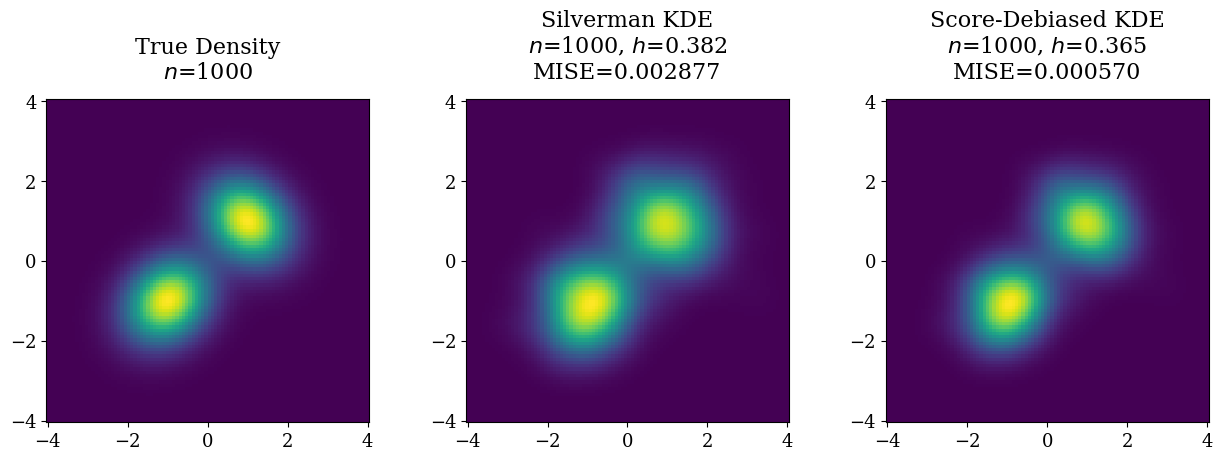}
  \vspace{-.2cm}
  \caption{
    Comparison of a true 2D mixture of Gaussians density vs. the Silverman method and our SD-KDE method using the true score. Given the oracle score function, SD-KDE outperforms Silverman in MISE by nearly an order of magnitude.
  }
  \label{fig:2d_mog}
\end{figure}

In Figure~\ref{fig:2d_mog}, on a mixture of Gaussians ground-truth density, we compare the Silverman method with SD-KDE. 

\section{MNIST Dataset Image}
The following figure depicts the ordering of generated images based on estimated probability density values. 
\begin{figure}[h]
  \centering
  \includegraphics[width=0.4\textwidth]{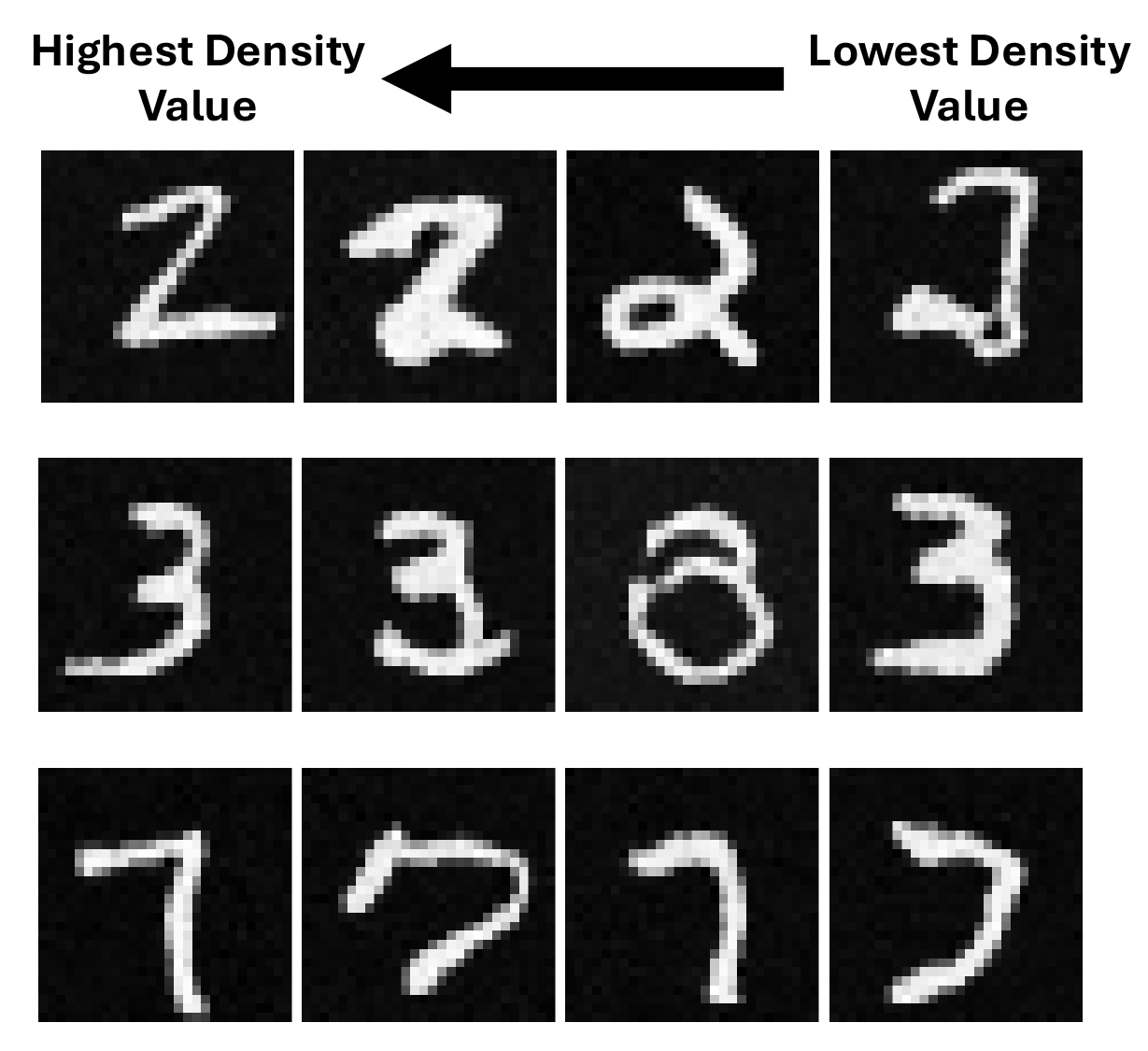}
  \caption{%
     Generated MNIST images of digits 2, 3, and 7 are displayed in descending order of estimated probability density as determined by score-based KDE. The ordering illustrates that images with higher probability density estimates exhibit more realistic features.
  }
  \label{fig:numbers}
\end{figure}

\end{document}